\documentclass[conference,letterpaper]{IEEEtran}

\usepackage[utf8]{inputenc} 
\usepackage[T1]{fontenc}
\usepackage{url}
\usepackage{ifthen}
\usepackage{cite}
\usepackage[cmex10]{amsmath} 

\usepackage{microtype}
\usepackage{graphicx}
\usepackage{subfigure}
\usepackage{booktabs,pifont} 

\usepackage{empheq}

\allowdisplaybreaks

\usepackage{pgfplots}
\pgfplotsset{compat=1.12}
\usetikzlibrary{shapes,arrows}
\usetikzlibrary{positioning}
\usepackage{tikz}
\usetikzlibrary{positioning,chains,fit,shapes,calc}
\usepackage{amsmath,bm,times}
\usetikzlibrary{calc}
\usepackage{dsfont}
\usepackage{cuted}
\usepackage{caption}

\usepackage{mathtools}

\usepackage{amsthm}
\usepackage{comment}
\usepackage{enumitem}
\usepackage{arydshln}
\usepackage{cite}
\usepackage{multirow}
\usepackage{calc}
\usepackage{blkarray}
\usepackage{subcaption}
\usepackage{url}
\theoremstyle{definition}

\usepackage{graphicx}
\usepackage{dblfloatfix}
\usetikzlibrary{decorations.pathreplacing}
\usepackage{blindtext, graphicx, amsfonts,
	amssymb,multirow,epstopdf}
\def\BibTeX{{\rm B\kern-.05em{\sc i\kern-.025em b}\kern-.08em
    T\kern-.1667em\lower.7ex\hbox{E}\kern-.125emX}}

\makeatletter
\renewcommand*\env@matrix[1][*\c@MaxMatrixCols c]{%
  \hskip -\arraycolsep
  \let\@ifnextchar\new@ifnextchar
  \array{#1}}
\makeatother
\usepackage[linesnumbered,ruled]{algorithm2e}

\usepackage[most]{tcolorbox}
\usepackage{xcolor}

\newcommand{\rqbox}[1]{%
\begin{tcolorbox}[
    colback=green!25,
    colframe=black!50,
    boxrule=0.4pt,
    arc=1.5pt,
    left=5pt,
    right=5pt,
    top=4pt,
    bottom=4pt,
    boxsep=0pt,
    before skip=6pt,
    after skip=6pt
]
#1
\end{tcolorbox}
}

\begin{document}
\title{Preventing Error Propagation in Multi-Agent AI \\ through Runtime Monitoring} 

\author{%
  \IEEEauthorblockN{Shahnewaz Karim Sakib}
  \IEEEauthorblockA{
                    University of Tennessee at Chattanooga, TN  37403, USA\\
                    Email: shahnewazkarim-sakib@utc.edu}
  \and
  \IEEEauthorblockN{Anindya Bijoy Das}
  \IEEEauthorblockA{
                    The University of Akron, OH 44325, USA\\
                    Email: adas@uakron.edu}
}


\maketitle

\begin{abstract}
Multi-agent AI systems can improve answer selection by allowing different language models to exchange reasoning traces, revise initial predictions, and support a final decision. However, such communication may also introduce reliability risks: reasoning from one agent can correct another agent’s mistake, but it can also mislead an agent that was initially correct. This paper studies reliable multi-agent AI communication through reasoning exchange and runtime answer revision. We develop a framework in which agents first answer multiple-choice questions independently, then share reasoning traces and revise their decisions. We conduct numerical experiments where we evaluate whether this process improves accuracy, produces more positive than negative answer transitions, and remains effective across domains such as cybersecurity, networking, and general knowledge. The results help identify when multi-agent reasoning improves reliability and when it may propagate errors.
\end{abstract}


\section{Introduction}
\label{sec:intro}

Large language model (LLM)-based systems are increasingly moving beyond single-agent question answering toward multi-agent workflows, where multiple agents communicate, retrieve information, critique intermediate outputs, and jointly produce final responses \cite{yao2022react, schick2023toolformer, li2023camel}. In such systems, one agent may serve as an answer generator, another may act as a verifier, while another may challenge unsupported assumptions or search for missing evidence \cite{li2023camel, wu2024autogen, qian2024chatdev}. This architecture can improve task decomposition and robustness, especially for complex problems requiring evidence gathering, reasoning, or tool use \cite{schick2023toolformer, liu2024agentbench}. However, it also introduces new reliability risks because the final answer is no longer produced by a single isolated model. Instead, it emerges from a sequence of messages, tool calls, retrieved evidences,  critiques, and revisions.

\begin{figure}[t]
    \centering
    \includegraphics[width=0.45\textwidth]{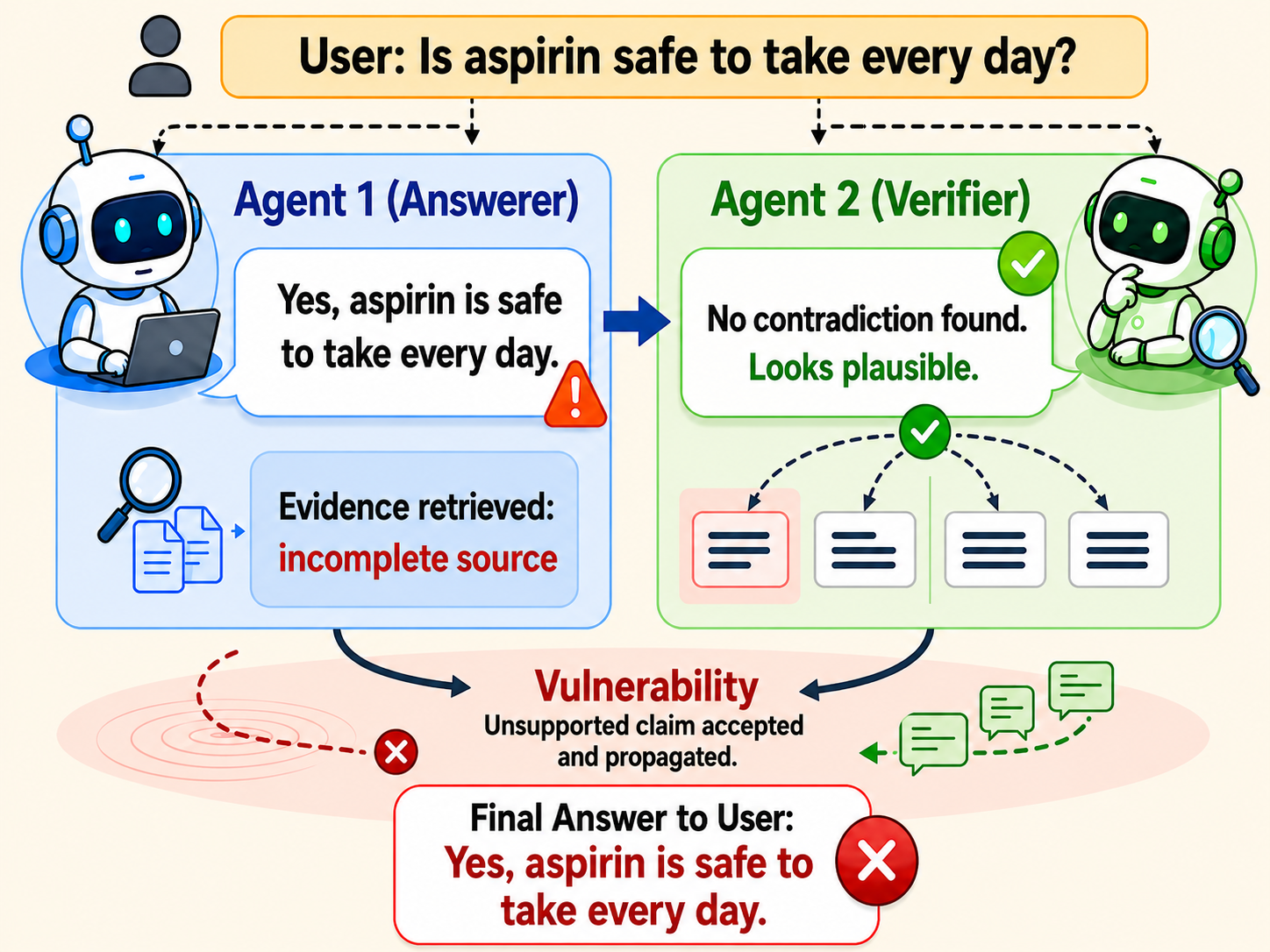}
    \vspace{-0.07 in}
    \caption{\footnotesize Example vulnerability in two-agent communication. Agent 1 generates a confident answer from incomplete evidence, and Agent 2 accepts it after a superficial verification step, allowing an unsupported claim to propagate to the final response. The fact is that Aspirin is not generally recommended for daily use unless prescribed or advised by a healthcare professional.}
    \vspace{-0.15 in}
    \label{fig:motivatingexample}
\end{figure}

\vspace{0.04 in}
A key challenge in multi-agent communication is that unreliable information can propagate across agents before the final answer is produced \cite{park2023generative}. For example, as shown in Fig. \ref{fig:motivatingexample}, an answering agent may retrieve incomplete evidence but still generate a confident response. A verifier agent may fail to identify the weak evidence support, or a challenger agent may focus on a secondary issue while missing the main unsupported claim. As a result, an initially unfaithful statement can become embedded in the shared communication context and later be treated as a valid premise by other agents \cite{manakul2023selfcheckgpt,lin2022truthfulqa}. This is particularly concerning under distribution shift, where user questions differ from the agents’ training or evaluation settings, or when external tools fail to retrieve up-to-date evidence \cite{lewis2020retrieval}. In these cases, the system may not only make an error, but may also amplify the error through interaction.

\vspace{0.04 in}
Most existing reliability mechanisms focus on evaluating the final answer, applying generic self-reflection, retrying a tool call, or using another model as a judge \cite{yuan2024r, zhuge2025agent}. While these strategies are useful, they may be insufficient for multi-agent systems because failures often emerge gradually during the communication process. A final-output check may detect that an answer is unsupported, but it does not necessarily reveal when the failure began, which agent contributed to it, whether tool use was unreliable, or what type of correction would be most effective \cite{shinn2023reflexion}. Therefore, trustworthy multi-agent communication requires runtime indicators that monitor the interaction before finalization. Such indicators should capture not only the final response quality, but also the reliability of intermediate communication, evidence use, disagreement, revision behavior, and confidence calibration \cite{wang2022self}.

This paper studies reliability in multi-agent AI systems through the lens of reasoning exchange, runtime detection, correction, and domain variation. We develop a multi-agent framework where language models first answer independently, then share reasoning traces, revise their answers, and optionally pass the revised outputs to an independent judging step. This setup allows us to examine whether reasoning from one agent can help another agent detect an unreliable initial answer and move toward a better decision. We then study runtime correction by measuring whether reasoning exchange fixes more initially wrong answers than it disrupts initially correct ones. Finally, we evaluate the framework across multiple domains, since reasoning quality and model reliability may vary between areas such as cybersecurity, networking, and machine learning. 

\section{Background and Summary of Contributions}
\label{sec:background}
This section reviews prior work on multi-agent LLM systems, tool-augmented reasoning, hallucination detection, and correction mechanisms. We first discuss how recent agent frameworks enable collaborative reasoning and tool use, then identify why existing verification methods remain insufficient for runtime reliability in multi-agent communication. Finally, we summarize the main contributions of this paper.

\subsection{Background and Motivations}
\label{sec:backgroundandmotivations}

In this section, we review prior work on LLM-based agents, tool use, hallucination detection, and verification, and discuss why runtime reliability and error propagation remain underexplored in multi-agent communication.

\subsubsection{Multi-agent communication and tool-augmented reasoning}
\label{sec:backreas}
LLM agents increasingly combine language generation with actions such as search, retrieval, code execution, and tool calls. The work in \cite{yao2022react} (ReAct) introduced an interleaved reasoning-and-acting paradigm, where models generate reasoning traces and external actions to improve decision making. Toolformer, proposed in \cite{schick2023toolformer}, further showed that language models can learn to call external APIs for tasks such as calculation, retrieval, and translation. Retrieval-augmented generation (RAG) also demonstrated that explicit non-parametric memory can improve factuality in knowledge-intensive tasks \cite{lewis2020retrieval}. These works move LLMs from closed-form text generation toward interactive systems that gather external information before answering. However, tool use also introduces new failure modes: retrieval may be incomplete, sources may be stale, and agents may incorrectly treat weak evidence as sufficient.

Multi-agent frameworks extend this idea by allowing several LLM-based agents to communicate, divide roles, and coordinate toward a final output. CAMEL studied role-playing communicative agents for autonomous cooperation \cite{li2023camel}, while AutoGen introduced a flexible framework for building applications through multi-agent conversation \cite{wu2024autogen}. AgentBench provided environments for evaluating LLMs as agents across interactive tasks \cite{liu2024agentbench}, and ChatDev demonstrated how specialized agents can collaborate through structured communication in software development \cite{qian2024chatdev}. These systems show the potential of multi-agent communication, but they also reveal a key risk: when one agent introduces an unsupported claim, later agents may accept, polish, or amplify it. Thus, reliability is not only a property of individual agents, but also of the communication process among them.

\subsubsection{Hallucination detection, verification, and correction}
\label{sec:backhal}
A large body of work studies hallucination, truthfulness, and answer verification in LLMs. TruthfulQA showed that models may reproduce common falsehoods even when they appear fluent and confident \cite{lin2022truthfulqa}. SelfCheckGPT proposed a black-box hallucination detection method based on consistency across sampled responses \cite{manakul2023selfcheckgpt}. Self-consistency improves reasoning by aggregating multiple reasoning paths \cite{wang2022self}, while Reflexion allows language agents to improve through verbal feedback stored in memory \cite{shinn2023reflexion}. These works motivate reliability mechanisms that go beyond a single generated response. However, most of them still operate at the level of final answers, repeated samples, or post-hoc reflection. They do not fully address failures that emerge through communication, such as one agent trusting another agent’s weak evidence or repeated claims being mistaken for independent support.
\vspace{0.07 in}

Recent correction methods also use critique, debate, retrieval, or verifier models to reduce hallucinations and improve factuality. Constitutional AI showed how AI feedback can guide self-critique and revision \cite{bai2022constitutional}, while multi-agent debate improves factuality and reasoning by allowing multiple model instances to compare and revise their responses through discussion \cite{du2024improving}. Retrieval-based revision can also edit unsupported claims using external evidence \cite{gao2023rarr}. However, these methods often rely on generic retry, reflection, debate, or model-judging strategies. In multi-agent communication, different failures may require different interventions, such as query rewriting for failed retrieval, cross-agent comparison for disagreement, constrained answering for weak evidence, or clarification for ambiguous intent. Our work shifts the focus from post-hoc verification to runtime detection and failure-specific correction, while studying how exchanged reasoning can both correct and propagate errors \cite{madaan2023self}.

\subsection{Summary of Contributions}
\label{sec:soc}
The contributions of this work are summarized below.
\begin{itemize}
    \item We develop a multi-agent communication setup in which language models first answer independently, then exchange reasoning traces and revise their decisions. This allows us to study how reasoning from one agent influences another agent's answer selection.

    \item We propose a runtime detection mechanism based on reasoning-trace comparison. By measuring how close or different the agents' reasoning traces are, we identify cases where communication may reveal uncertainty, disagreement, or possible unreliability before final selection.

    \item We design a reasoning-guided correction mechanism that combines support from multiple agents' reasoning traces rather than relying only on initial answer choices. The correction step selects the answer that is better supported by the exchanged reasoning.

    \item We conduct numerical experiments across different domains to evaluate whether reasoning exchange improves accuracy, corrects more initially wrong answers than it disrupts correct ones, and remains effective under domain-dependent reasoning variation.
\end{itemize}

\section{Proposed Approach}
\label{sec:approach}

This section presents our multi-agent framework for studying runtime reliability in language-model communication. We first prepare a controlled multi-agent setting where different agents independently answer a question, exchange reasoning traces, and revise their decisions. We then use this interaction to study runtime error detection, error correction, and the research questions (RQ) on whether agent communication improves accuracy, corrects more mistakes than it introduces, and remains effective across domains.

\subsection{Preparing the Multi-Agent System}
\label{sec:prepagent}
We consider a multi-agent system composed of K collaborative, reasoning-enabled language-model agents,
\begin{align*}
    \mathcal{A}=\{A_1,A_2,\ldots,A_K\}.    
\end{align*}
Given a multiple-choice question \(q\) with candidate options \(\mathcal{O}=\{o_1,o_2,\ldots,o_M\}\), each agent first produces an independent answer and a reasoning trace. The answer represents the option selected by the agent, while the reasoning trace contains the intermediate explanation used to justify that choice. This independent stage is important because it provides a baseline for evaluating whether communication among agents improves or harms the original predictions. It also allows us to observe whether different agents make the same mistake, disagree with each other, or provide complementary reasoning that may help later correction.

After the independent stage, the agents communicate by sharing their reasoning traces and revising their answers. For an agent \(A_i\), its initial output is written as
\begin{equation}
    (y_i,r_i)=A_i(q,\mathcal{O}),
\end{equation}
where \(y_i\in\mathcal{O}\) is the selected answer and \(r_i\) is the generated reasoning trace. During communication, another agent’s reasoning can be used as additional context for revision. This setup allows us to study error propagation and correction at runtime: an incorrect answer may be reinforced if another agent accepts weak reasoning, or corrected if the shared reasoning reveals missing evidence or a better explanation. Therefore, the multi-agent system provides a controlled environment for examining when communication improves reliability and when it introduces new failures.

\subsection{Runtime Error Detection}

The runtime error detection stage examines whether the reasoning traces produced by two agents are aligned or substantially different before the final answer is selected. The intuition is that two agents may choose different options because their reasoning focuses on different evidence, assumptions, or elimination steps. Similarly, even when the selected answers are the same, their reasoning traces may differ in quality or support. Therefore, instead of only comparing final options, we compare the reasoning traces to detect whether the agents are moving toward a reliable or potentially unstable decision.


Let \(r_i\) and \(r_j\) denote the reasoning traces generated by agents \(A_i\) and \(A_j\), respectively. We define the reasoning distance as
\begin{equation}
    d_{ij}=1-\mathrm{sim}(r_i,r_j),
\end{equation}
where \(\mathrm{sim}(r_i,r_j)\in[0,1]\) measures semantic similarity. A small \(d_{ij}\) means the agents use similar reasoning, while a large \(d_{ij}\) suggests different or potentially conflicting rationales. This distance helps identify when reasoning exchange may be useful: complementary traces may correct an error, whereas misleading traces may introduce one. This detection step is directly connected to our first research question. If reasoning traces contain useful information beyond the selected answer, then allowing one model to use another model’s reasoning should improve answer selection compared with isolated predictions. We therefore evaluate whether reasoning combination improves performance using the same metrics used throughout the paper: accuracy, positive impact, and negative impact.

\rqbox{\textbf{RQ1}: Does combining reasoning traces from two language models improve multiple-choice answer accuracy compared with their original predictions?}

\subsection{Runtime Error Correction}

The runtime correction stage uses exchanged reasoning traces to revise unreliable answers before the final decision. Correction should depend not only on whether two agents select the same option, but also on whether their reasoning sufficiently supports it. If one agent selects a wrong answer but provides useful evidence, another agent may use that trace to move toward the correct option. Conversely, weak or misleading reasoning may disturb an initially correct answer. Thus, correction is framed as reasoning-guided revision rather than majority voting.

To capture this idea, we define a reasoning-supported correction score for each candidate option \(o\in\mathcal{O}\):
\begin{equation}
\label{eq:s(o)}
    S(o)=\lambda_1\,\phi(o,r_1)+\lambda_2\,\phi(o,r_2)
    -\lambda_3\,\delta(r_1,r_2,o),
\end{equation}
where \(\phi(o,r_i)\) measures how strongly reasoning trace \(r_i\) supports option \(o\), and \(\delta(r_1,r_2,o)\) measures unresolved conflict between the two reasoning traces with respect to that option. The weights \(\lambda_1,\lambda_2,\lambda_3\) control the influence of each reasoning trace and the penalty for contradiction. The corrected answer is then selected as the option with the strongest combined reasoning support, rather than the option that was initially chosen most confidently.

This formulation motivates the need to distinguish useful correction from harmful revision. Accuracy alone does not reveal whether improvement comes from fixing wrong answers or from a mixture of corrections and newly introduced errors. Therefore, after applying reasoning-guided correction, we separately analyze whether the process changes initially wrong answers into correct ones more often than it changes initially correct answers into incorrect ones.

\rqbox{\textbf{RQ2}: Does reasoning combination correct more initially wrong answers than it changes initially correct answers into incorrect ones?}

\subsection{Domain-Aware Reasoning Variation}
\label{sec:domainreasoning}


The effectiveness of reasoning combination may vary across domains because the same correction score can behave differently depending on the type of reasoning required. For example, cybersecurity questions may require reasoning about vulnerabilities, attack behavior, and adversarial actions, while networking questions may require protocol logic, routing behavior, and performance trade-offs. Therefore, even with the same correction rule in \eqref{eq:s(o)}, the support term \(\phi(o,r_i)\) and the conflict term \(\delta(r_1,r_2,o)\) may vary substantially across domains. In one domain, two reasoning traces may provide complementary support for the correct option, while in another domain, one trace may introduce misleading evidence or increase unresolved conflict.

Thus, rather than introducing a separate correction rule for each domain, we use the same reasoning-supported score \(S(o)\) and analyze how its components behave across domains. A domain where both agents provide consistent support for the correct option should lead to a stronger combined score, while a domain with misleading or contradictory reasoning may increase the conflict penalty and reduce the benefit of correction. This allows us to study whether reasoning-guided correction is generally reliable or whether its effectiveness depends on domain-specific reasoning quality, such as cybersecurity versus networking or machine learning.

This motivates our domain-level analysis. If the support and conflict components of reasoning traces vary across subject areas, then reasoning combination may not provide uniform gains across all domains. Therefore, we examine whether the observed benefits are consistent across the considered domains or whether some domains are more vulnerable to misleading or conflicting reasoning.

\rqbox{\textbf{RQ3}: To what extent does the effectiveness of reasoning combination vary across different subject domains?}
\vspace{0.07 in}

\section{Experimental Results}
\label{sec:numexp}
\subsection{Experimental Setup}

We evaluate the proposed framework using three domain-specific subsets from Open Quiz Commons: cybersecurity, machine learning, and networking \cite{openquizcommons}. Each subset contains MCQs with four answer choices and a corresponding ground-truth answer. For every domain, we use outputs from two base models, Phi3-14B \cite{phi3} and Gemma2-9B \cite{gemma2}, where each output includes the model's predicted reasoning for the same set of questions. These paired outputs allow us to examine whether combining reasoning traces from different models can improve answer selection. All answers are normalized to a common multiple-choice label format before evaluation.

\begin{figure}[t]
    \centering
    \includegraphics[width=0.99\linewidth]{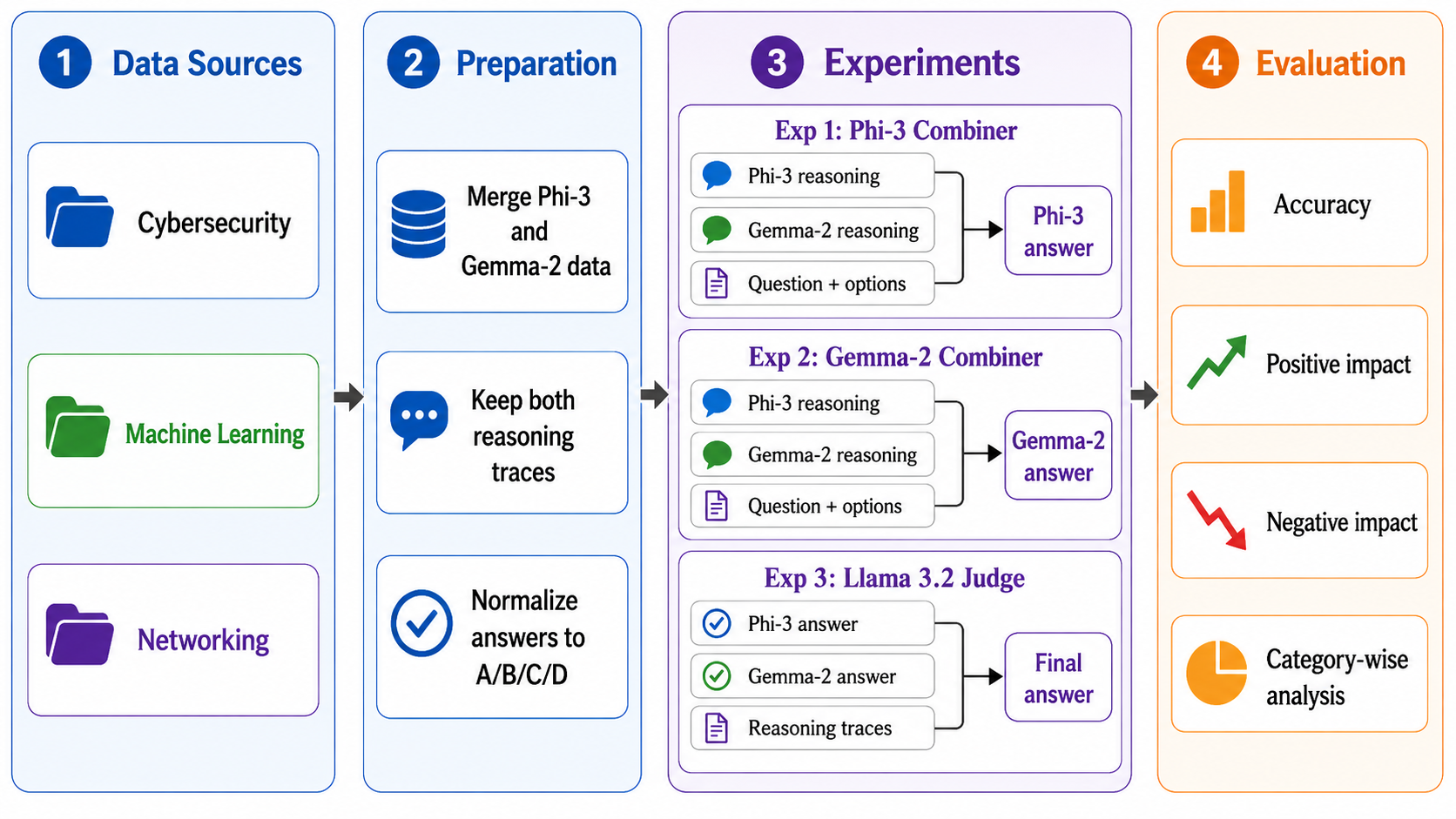}
    \vspace{-0.15 in}
    \caption{\small Overview of the experimental pipeline. The process begins with domain-specific datasets, followed by preprocessing, reasoning-combination experiments, and independent judging.}
    \vspace{-0.1 in}
    \label{fig:experimental_pipeline}
\end{figure}

Fig.~\ref{fig:experimental_pipeline} illustrates the overall experimental pipeline. For each Open Quiz Commons domain subset, the corresponding Phi-3 and Gemma-2 outputs are loaded and aligned using the shared question and answer options. After preprocessing, both reasoning traces are retained and used in three experimental stages. First, Phi-3 combines its own reasoning with Gemma-2's reasoning to produce a revised answer. Second, Gemma-2 performs the same reasoning-combination process. Finally, Llama 3.2 \cite{meta2024llama32} acts as an independent judge by considering the two combined answers and the original reasoning traces to produce the final answer. The resulting predictions are evaluated using accuracy and impact-based metrics for each domain.

\subsection{Evaluation Metrics}

We use three metrics to evaluate performance: accuracy, positive impact, and negative impact. All answers are converted to the same label format, $\{A, B, C, D\}$, before comparison with the ground truth. We report these metrics both overall and by subject domain.

\paragraph{Accuracy.}
Accuracy measures the proportion of examples for which the predicted answer matches the ground-truth answer. Given a dataset $D$ containing $N$ examples, let $y_i$ denote the ground-truth label for example $i$, and let $\hat{y}_i$ denote the predicted label. Accuracy is defined as:
\vspace{-0.07 in}
\[
\mathrm{Accuracy} = \frac{1}{N} \sum_{i=1}^{N} \mathbb{I}(\hat{y}_i = y_i),
\]
\vspace{-0.1 in}

\noindent where $\mathbb{I}(\cdot)$ is an indicator function that returns 1 when the condition is true and 0 otherwise.

\paragraph{Positive Impact.}
Positive impact captures cases where the process improves an initially incorrect prediction. Let $b_i$ denote the answer before the combination or judging step, and let $a_i$ denote the answer after that step. It is defined as:
\vspace{-0.07 in}
\[
\mathrm{Positive\ Impact} = \sum_{i=1}^{N} \mathbb{I}(b_i \neq y_i \land a_i = y_i).
\]
\vspace{-0.1 in}

\noindent This metric measures how often the method corrects an error.

\paragraph{Negative Impact.}
Negative impact captures cases where the process changes an initially correct prediction into an incorrect one. It is defined as:
\vspace{-0.05 in}
\[
\mathrm{Negative\ Impact} = \sum_{i=1}^{N} \mathbb{I}(b_i = y_i \land a_i \neq y_i).
\]
\vspace{-0.1 in}

\noindent This metric measures how often the method introduces a new error by disrupting an answer that was already correct before reasoning combination.

\subsection{Findings by Research Question}

\begin{figure}[t]
    \centering
    \includegraphics[width=0.9\linewidth]{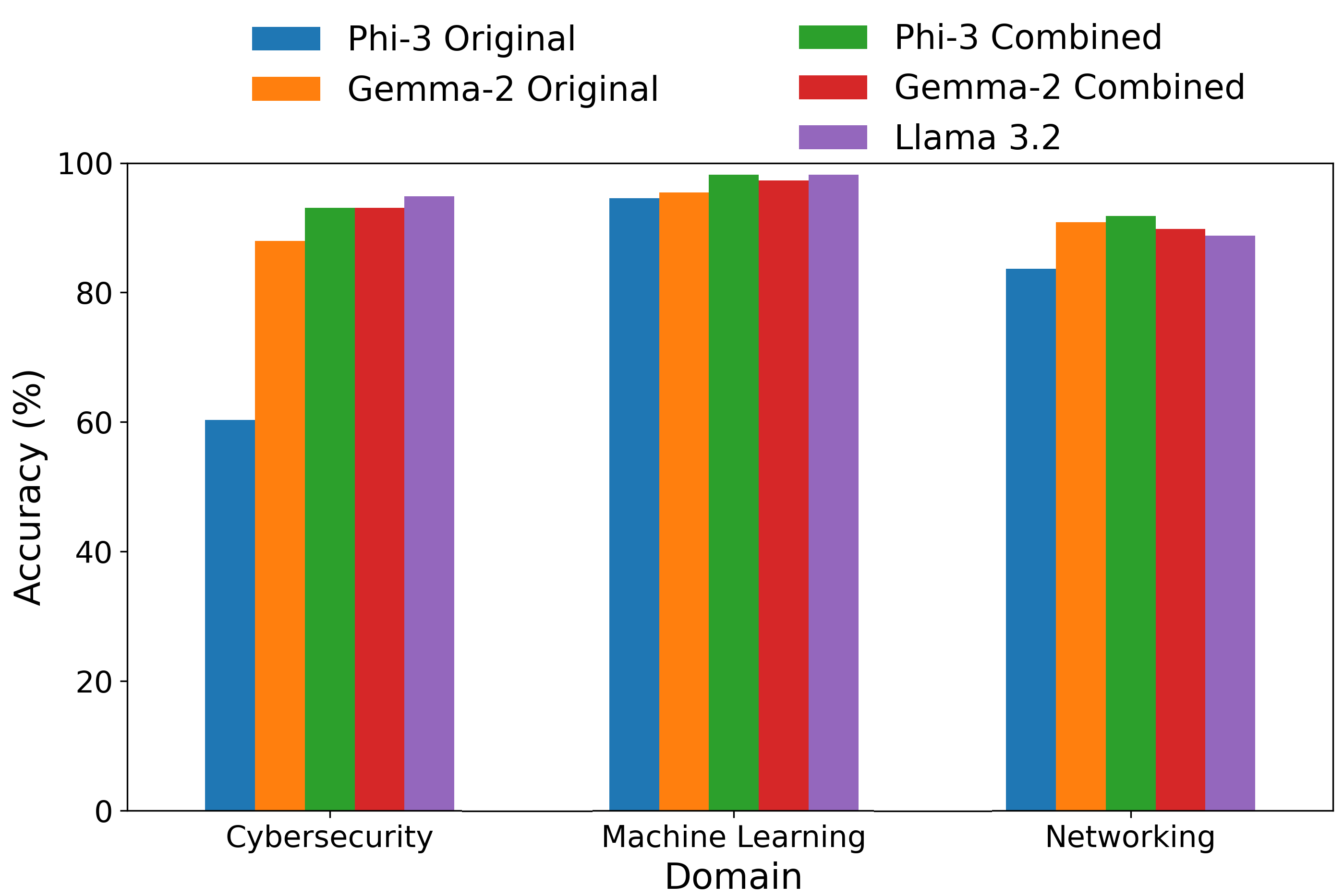}
        \vspace{-0.1 in}
        \caption{\small Domain-wise accuracy comparison across original predictions, reasoning-combined predictions, and LLaMA 3.2 predictions.}
    \label{fig:accuracy_comparison}
    \vspace{-0.1 in}
\end{figure}

\paragraph{Does reasoning combination improve accuracy?}
The results show that reasoning combination generally improves answer accuracy, although the improvement is not uniform across all model-domain settings. As shown in Table~\ref{tab:domain_performance} and Figure~\ref{fig:accuracy_comparison}, the largest gain occurs in the cybersecurity domain, where Phi-3 improves from 60.34\% to 93.10\% after incorporating Gemma-2's reasoning. Gemma-2 also improves in the same domain, increasing from 87.93\% to 93.10\%. In ML, the original accuracies are already high, so the gains are smaller but still positive for both models. Networking shows a more mixed pattern: Phi-3 improves from 83.67\% to 91.84\%, while Gemma-2 slightly decreases from 90.82\% to 89.80\%. This indicates that reasoning combination can improve answer selection when the original model has greater room for correction, but it does not guarantee improvement in every setting.

\begin{table*}[t]
\centering
\caption{\small Accuracy comparison across domains for original predictions, reasoning-combined predictions, and final Llama 3.2 predictions.}
\label{tab:domain_performance}
\begin{tabular}{lccccc}
\toprule
Domain & Phi-3 Original & Gemma-2 Original & Combined by Phi-3  & Combined by Gemma-2 & Combined by Llama 3.2  \\
\midrule
Cybersecurity & 60.34\% & 87.93\% & 93.10\% & 93.10\% & 94.83\% \\
Machine Learning & 94.55\% & 95.45\% & 98.18\% & 97.27\% & 98.18\% \\
Networking & 83.67\% & 90.82\% & 91.84\% & 89.80\% & 88.78\% \\
\bottomrule
\end{tabular}
\end{table*}

\begin{figure*}[t]
    \centering

    \begin{minipage}[t]{0.32\textwidth}
        \centering
        \includegraphics[width=\linewidth]{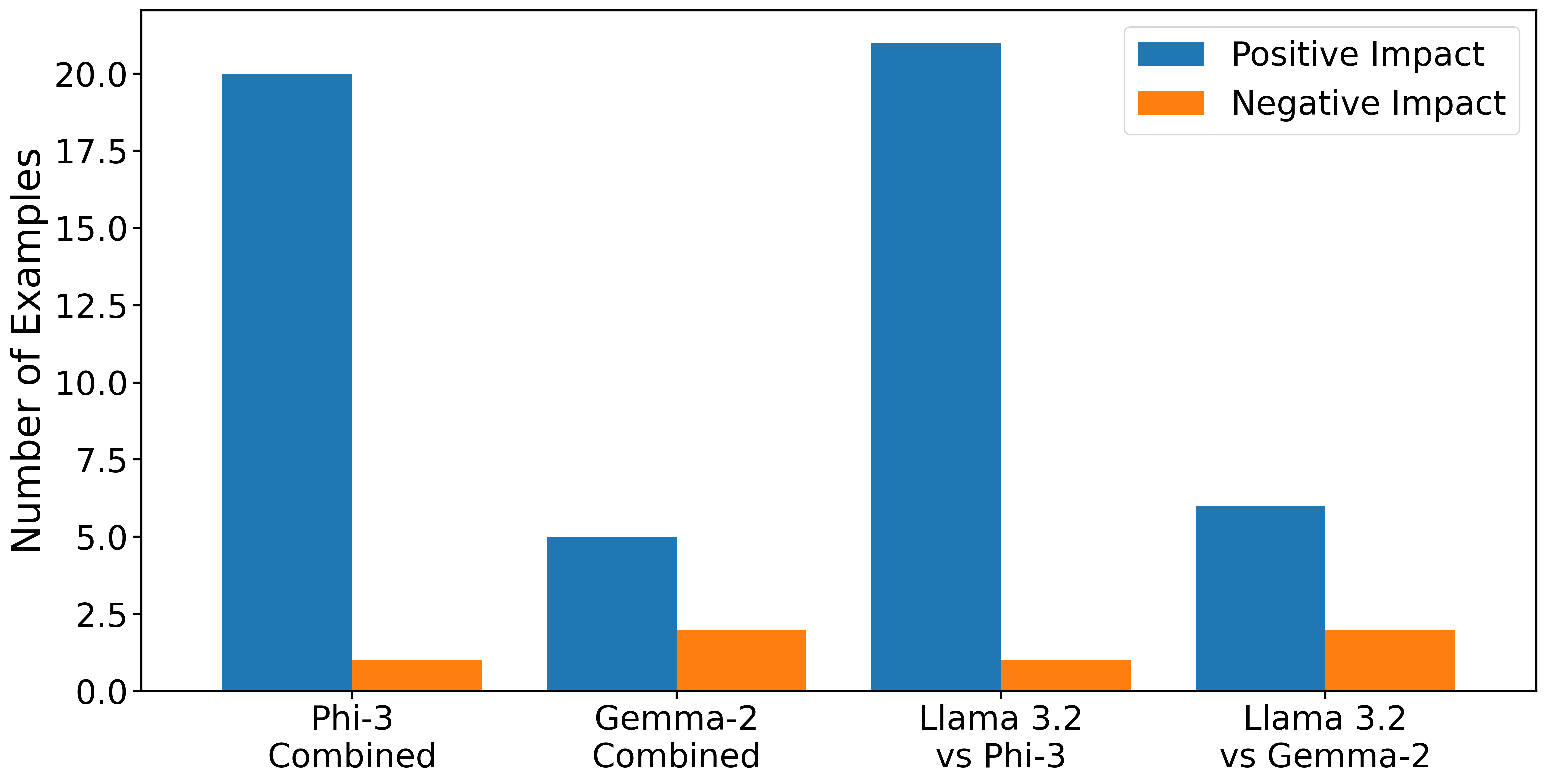}
        \caption*{\small (a) Cybersecurity}
    \end{minipage}
    \hfill
    \begin{minipage}[t]{0.32\textwidth}
        \centering
        \includegraphics[width=\linewidth]{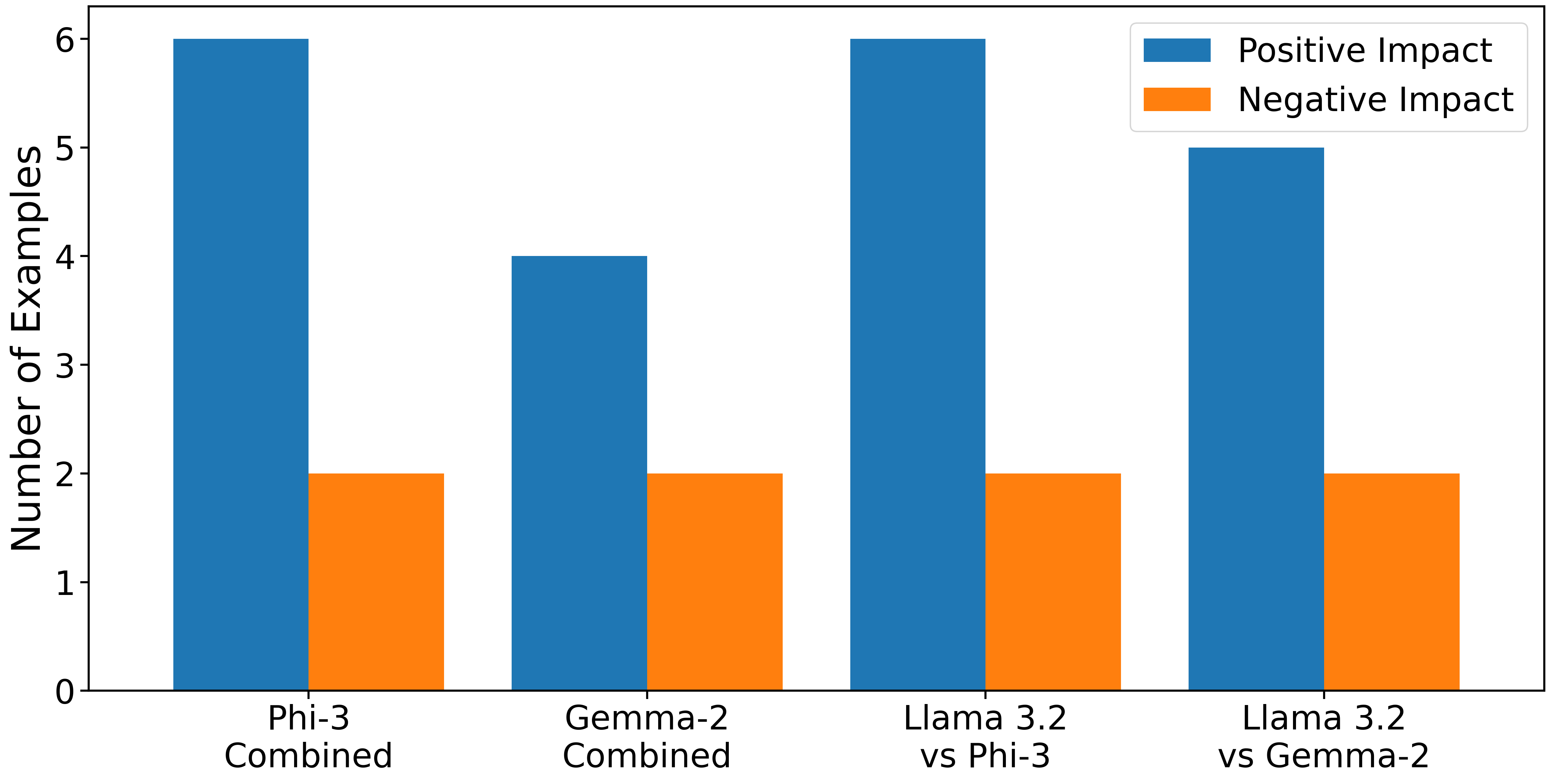}
        \caption*{\small (b) Machine Learning}
    \end{minipage}
    \hfill
    \begin{minipage}[t]{0.32\textwidth}
        \centering
        \includegraphics[width=\linewidth]{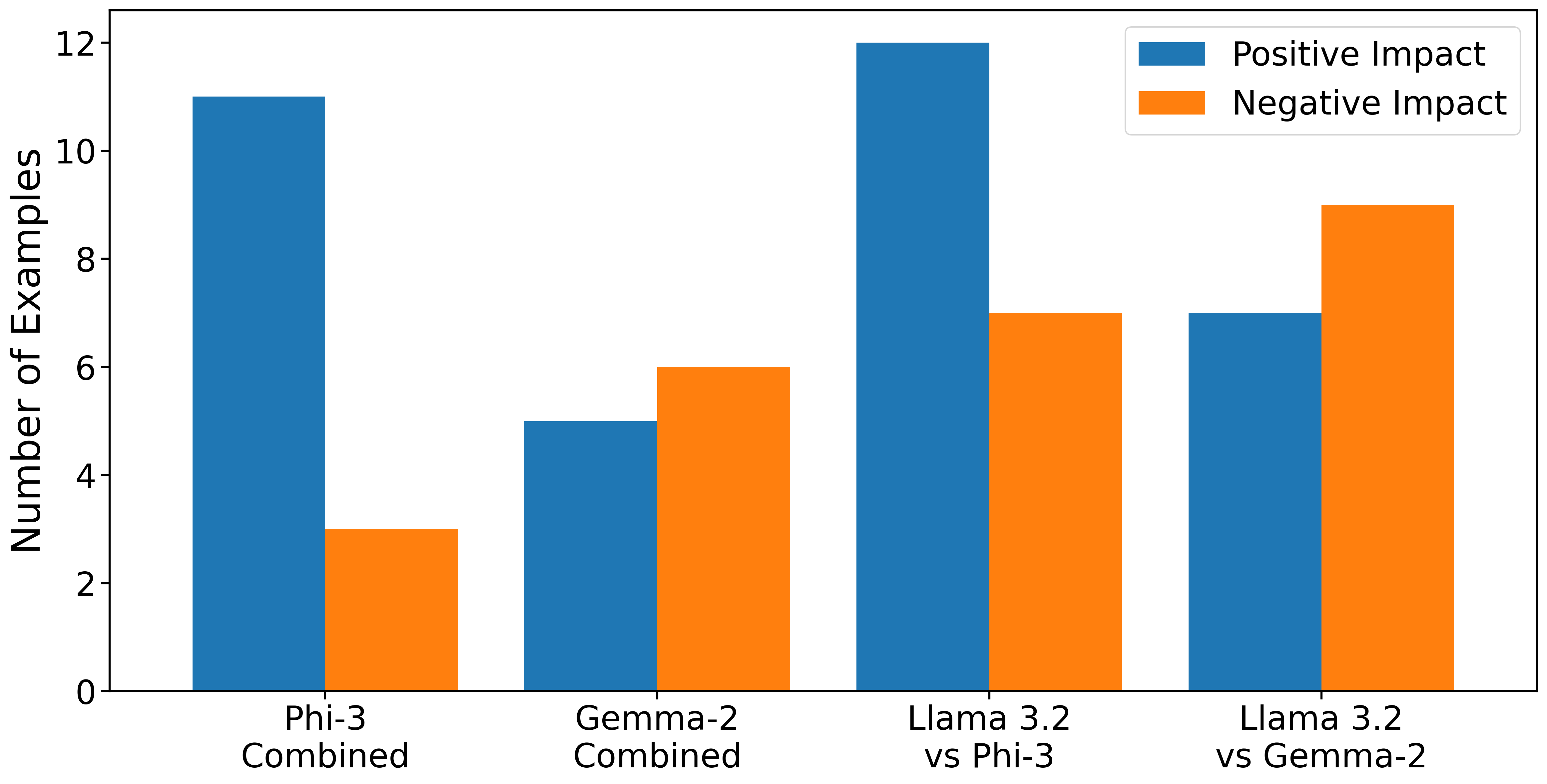}
        \caption*{\small (c) Networking}
    \end{minipage}
    \vspace{-0.05in}
    \caption{\small Category-wise positive and negative impact across reasoning-combination and Llama 3.2 judging conditions. Positive impact indicates where an initially incorrect answer becomes correct, while negative impact indicates cases where an initially correct answer becomes incorrect.}
        \vspace{-0.15 in}
    \label{fig:category_impact}
\end{figure*}

\paragraph{Does it correct more errors than it introduces?}
The impact analysis indicates that reasoning combination usually corrects more errors than it introduces, but this behavior depends on the condition being evaluated. Table~\ref{tab:impact_analysis} and Figure~\ref{fig:category_impact} show that the strongest positive impact appears in cybersecurity. For example, Phi-3 Combined corrects 20 initially wrong answers while introducing only 1 new error, producing a net gain of 19. A similar pattern is observed for Llama 3.2 when evaluated relative to Phi-3. In machine learning, the net gains are smaller because the original predictions are already strong, leaving fewer incorrect answers available to correct. Networking is less stable: Phi-3 Combined still produces a positive net impact, but Gemma-2 Combined and Llama 3.2 relative to Gemma-2 introduce slightly more errors than corrections. Thus, reasoning combination is often beneficial, but the impact metrics reveal cases where additional reasoning can also mislead the final decision.

\begin{table*}[t]
\centering
\caption{\small Domain-wise impact analysis showing how often each condition corrects initially wrong answers or changes initially correct answers into incorrect ones.}
\label{tab:impact_analysis}
\begin{tabular}{cccccccc}
\toprule
Domain & Condition & Positive Count & Negative Count & Net Count & Positive Rate & Negative Rate & Net Rate \\
\midrule
\multirow{4}{*}{Cybersecurity}
& Phi-3 Combined & 20 & 1 & 19 & 34.48\% & 1.72\% & {\bf 32.76\% }\\
& Gemma-2 Combined & 5 & 2 & 3 & 8.62\% & 3.45\% & {\bf 5.17\% }\\
& Llama 3.2 vs Phi-3 & 21 & 1 & 20 & 36.21\% & 1.72\% & {\bf 34.48\% }\\
& Llama 3.2 vs Gemma-2 & 6 & 2 & 4 & 10.34\% & 3.45\% & {\bf 6.90\% }\\
\midrule
\multirow{4}{*}{Machine Learning}
& Phi-3 Combined & 6 & 2 & 4 & 5.45\% & 1.82\% & 3.64\% \\
& Gemma-2 Combined & 4 & 2 & 2 & 3.64\% & 1.82\% & 1.82\% \\
& Llama 3.2 vs Phi-3 & 6 & 2 & 4 & 5.45\% & 1.82\% & {\bf 3.64\%} \\
& Llama 3.2 vs Gemma-2 & 5 & 2 & 3 & 4.55\% & 1.82\% & 2.73\% \\
\midrule
\multirow{4}{*}{Networking}
& Phi-3 Combined & 11 & 3 & 8 & 11.22\% & 3.06\% & {\bf 8.16\%} \\
& Gemma-2 Combined & 5 & 6 & -1 & 5.10\% & 6.12\% & -1.02\% \\
& Llama 3.2 vs Phi-3 & 12 & 7 & 5 & 12.24\% & 7.14\% & {\bf 5.10\%} \\
& Llama 3.2 vs Gemma-2 & 7 & 9 & -2 & 7.14\% & 9.18\% & -2.04\% \\
\bottomrule
\end{tabular}
\end{table*}

\paragraph{Does the effect vary by domain?}
The results demonstrate clear domain-dependent behavior. Cybersecurity benefits the most from reasoning combination, both in terms of accuracy and positive impact. This suggests that the additional reasoning trace provides useful complementary information in a domain where one of the original models performs substantially weaker. Machine learning shows smaller improvements because both models begin with high baseline accuracy, making large gains less likely. Networking exhibits the most inconsistent behavior, with improvements for Phi-3 but weaker outcomes for Gemma-2 and Llama 3.2 relative to Gemma-2. Overall, the findings suggest that cross-model reasoning is most effective when the second trace corrects a weaker initial prediction, but its benefit decreases when the original prediction is reliable or the additional reasoning introduces conflicting evidence.


\section{Conclusion}
\label{sec:conclusion}
This paper studied multi-agent communication as a runtime reliability problem, focusing on whether reasoning traces from different language models can improve multiple-choice answer selection. We developed a framework where agents first answer independently, exchange reasoning traces, revise their decisions, and use a judging step for final selection. The results show how reasoning exchange can improve accuracy, produce positive corrections, but also introduce negative changes when misleading traces are accepted. The study further highlights that the benefit of multi-agent reasoning varies across domains, indicating that communication quality depends on both model behavior and subject-specific reasoning demands. 
Future work will extend this framework to larger agent teams with dynamic roles, where agents can be selected or assigned responsibilities based on task difficulty and reasoning disagreement, following recent multi-agent designs such as MetaGPT \cite{hong2024metagpt} and DyLAN \cite{liu2023dynamic}. Another direction is to develop domain-adaptive \cite{saunders2019domain}  weighting strategies that selectively combine each agent’s reasoning based on domain-specific reliability \cite{chen2024agentverse}.

\bibliographystyle{IEEEtran}
\bibliography{citations}


\end{document}